\documentclass{article}
\usepackage[utf8]{inputenc}
\usepackage{amsmath} 
\usepackage{amssymb} 
\usepackage{amsfonts} 
\usepackage{graphicx} 
\usepackage{hyperref} 
\usepackage{enumitem} 
\usepackage{float} 
\usepackage{booktabs} 
\usepackage{bm} 
\usepackage{cite}

\title{System-Level Uncertainty Quantification with Multiple Machine Learning Models: A Theoretical Framework}
\author{Xiaoping Du\\ \small School of Mechanical Engineering \\ \small Purdue University }
\date{\today} 

\begin{document}
	
	\maketitle
	
	\begin{abstract}
		ML models have errors when used for predictions. The errors are unknown but can be quantified by model uncertainty. When multiple ML models are trained using the same training points, their model uncertainties may be statistically dependent. In reality, model inputs are also random with input uncertainty. The effects of these types of uncertainty must be considered in decision-making and design. This study develops a theoretical framework that generates the joint distribution of multiple ML predictions given the joint distribution of model uncertainties and the joint distribution of model inputs. The strategy is to decouple the coupling between the two types of uncertainty and transform them as independent random variables. The framework lays a foundation for numerical algorithm development for various specific applications.
	\end{abstract}
	
	\section{Introduction}
	The increasing adoption of machine learning (ML) models in engineering analysis and design has enhanced decision-making processes by providing rapid predictions for complex systems \cite{Alam2024, panchal2019machine}. However, the inherent nature of ML predictions introduces critical challenges related to uncertainty. This section introduces the background and concepts of uncertainty in ML-based predictions, reviews relevant literature, and outlines the motivation and contributions of this paper.
	\subsection{Sources of Uncertainty in ML Predictions}
	In real-world engineering applications, two primary sources of uncertainty affect ML model predictions:
	\begin{itemize}
		\item \textbf{Model Uncertainty (Epistemic Uncertainty):} ML models are approximations of underlying physical phenomena or complex simulations. Their predictions are subject to inherent errors due to limited training data, model misspecification, or extrapolation beyond the sampled design space. This uncertainty, often termed model or epistemic uncertainty, reflects the lack of knowledge about the true system behavior. Many advanced ML methods, such as Gaussian Process (GP) regression \cite{rasmussen2006gaussian} and Bayesian Neural Networks \cite{neal2012bayesian}, are designed to provide not only point predictions but also probabilistic outcomes (e.g., mean and variance) that quantify this epistemic uncertainty.
		\item \textbf{Input Uncertainty (Aleatory Uncertainty):} Beyond the model itself, the input parameters to ML models are rarely deterministic in practical scenarios. Material properties, environmental conditions, manufacturing tolerances, and operational loads often exhibit inherent variability. This variability, known as input uncertainty, is aleatory in nature, representing the irreducible randomness.
	\end{itemize}
	
	\subsection{Coupled Uncertainty and its Consequences}
	A significant challenge arises because these two types of uncertainty are often coupled. Specifically, the distribution parameters of model uncertainty (e.g., the predictive mean and variance of a GP model) are functions of the random inputs. This means that the epistemic uncertainty itself varies depending on the aleatory uncertainty in the inputs. The presence of such coupled uncertainty has profound consequences for predictions, decision-making, and design:
	\begin{itemize}
		\item \textbf{Unreliable Predictions:} Ignoring or improperly handling coupled uncertainty can lead to predictions that are overly confident or inaccurate, potentially compromising the safety and reliability of engineered systems.
		\item \textbf{Suboptimal Decisions:} Engineering decisions, such as design optimization or risk assessment, rely on accurate probabilistic information. Coupled uncertainty can obscure the true risk picture, leading to suboptimal or even unsafe design choices.
		\item \textbf{Inefficient Designs:} Designs that do not account for the full spectrum of uncertainties may be either over-designed (leading to unnecessary cost) or under-designed (leading to potential failure).
	\end{itemize}
	Therefore, it is crucial to accurately estimate the probability distribution of a machine learning prediction by accounting for all sources of uncertainty. Existing approaches to handling coupled uncertainty often focus on predicting marginal distributions of individual model outputs and rely on the assumption that these outputs are independent.
	
	\section{System Uncertainty Quantification}
	
	\subsection{Dependent Model Uncertainties and System Uncertainty Quantification}
	System uncertainty quantification (UQ), such as system reliability analysis under input uncertainty, is inherently challenging. The difficulty increases significantly when multiple machine learning (ML) models are involved—particularly when their predictive uncertainties are statistically dependent. This dependence often arises when the models are trained concurrently on shared datasets. For example, in multi-output Gaussian Processes (e.g., Co-Kriging), the outputs of several GP models may follow a joint distribution, indicating statistical dependence among their predictive uncertainties. In such cases, a rigorous approach to system UQ is required to evaluate the combined effects of both input uncertainty and dependent model uncertainties on multiple ML predictions. The complexity of system UQ in this context arises from:
	\begin{itemize}
		\item The need to characterize and propagate the joint distribution of input uncertainty.
		\item The need to characterize and propagate the joint distribution of dependent model uncertainty.
		\item The coupling between input and model uncertainties, where distribution parameters of model uncertainty are functions of random inputs.
		\item The challenge of combining these dependent and coupled random variables to obtain the joint distribution of multiple ML predictions.
	\end{itemize}
	
	\subsection{Overview of this Study}
	This study addresses these complexities by developing a theoretical framework for system UQ. The key features are:
	\begin{itemize}
		\item We propose a strategy to explicitly decouple the inherent coupling between input uncertainty and model uncertainty, transforming them into independent random variables. This simplifies the subsequent propagation of uncertainty.
		\item We provide a detailed theoretical framework for generating the joint distribution of multiple ML predictions, given the joint distribution of model uncertainties (including dependent ones) and the joint distribution of model inputs.
		\item We present fundamental probabilistic transformations, including the conversion of correlated Gaussian variables into independent standard normal variables, which forms a cornerstone of the proposed framework.
		\item This framework lays a robust foundation for the development of numerical algorithms applicable to a wide range of specific engineering applications, enabling more reliable predictions and informed decision-making under complex uncertainty conditions.
	\end{itemize}
	
	We treat all uncertainty in an ML model as model uncertainty. It arises from both aleatory sources (such as data noise) and epistemic sources (such as model limitations). For engineering users, these two parts may not be separated. Once a model is trained and delivered, the user cannot reduce the aleatory uncertainty within it or retrain the model to address epistemic uncertainty. Therefore, the total predictive uncertainty is taken as a single source of model uncertainty. 
		
	\section{Methodology}
	
	\subsection{The Task}
	This section outlines the theoretical framework for propagating system uncertainty in ML model predictions. The primary task is to determine the joint distribution of multiple ML model outputs, considering both input and model uncertainties, particularly when model uncertainties are statistically dependent.
	
	\textbf{Given:}
	\begin{itemize}
		\item ML models with input vector $\mathbf{X} = (X_1, \dots, X_{n_X})^T$ and output vector $\mathbf{Y} = (Y_1, \dots, Y_{n_Y})^T$. Each ML model is defined as $Y_i = g_i(\mathbf{X})$, for $i=1,2,\dots,n_Y$.
		\item The joint distribution of model uncertainty for $\mathbf{Y}$, which is characterized by the conditional joint CDF of $\mathbf{Y}$ given $\mathbf{X}$, denoted as $F_{\mathbf{Y}|\mathbf{X}}(\mathbf{y}; \boldsymbol{\theta}_Y(\mathbf{X}))$. Here, $\boldsymbol{\theta}_Y(\mathbf{X})$ represents the parameters of this joint conditional distribution, which are functions of $\mathbf{X}$.
		\item The joint distribution of $\mathbf{X}$, characterized by its joint cumulative distribution function (CDF), $F_{\mathbf{X}}(\mathbf{x})$.
	\end{itemize}
	
	\textbf{Find:}
	\begin{itemize}
		\item The joint distribution of the output vector $\mathbf{Y}$, characterized by its joint CDF, $F_{\mathbf{Y}}(\mathbf{y})$.
	\end{itemize}
	
	This section first considers the scenario with only model uncertainty and then extends the results to include coupled uncertainty.
	
	\subsection{Case 1: No Input Uncertainty in \texorpdfstring{$\mathbf{X}$}{X} (X=x, deterministic)}
	This subsection considers the scenario where the input vector $\mathbf{X}$ is deterministic, denoted as $\mathbf{x}$. In this case, there is no input uncertainty, and the focus is solely on quantifying the effects of model uncertainty on ML model prediction at this fixed input $\mathbf{x}$.
	
	For each ML model output $Y_i$ in the vector $\mathbf{Y}$, its probabilistic response is characterized by a marginal distribution $F_{Y_i}(Y_i; \boldsymbol{\theta}_i)$. An auxiliary scalar random variable $Z_i$ for each $Y_i$ is used for the probability transform:
	\begin{equation}\label{eq:prob_integral_transform_Z_vector_epistemic}
		Z_i = F_{Y_i}(Y_i;\boldsymbol{\theta}_i(\mathbf{x})) \quad \text{for } i=1, \dots, n_Y
	\end{equation}
	where $\boldsymbol{\theta}_i(\mathbf{x})$ contains the distribution parameters. Using the auxiliary scalar random variable $Z_i$ has been applied in component reliability analysis \cite{Ditlevsen1982,Ditlevsen1996,Bjerager1990,rasmussen2006gaussian,Nannapaneni2016,Du_UQ_2024}. It is also the fundamental principle used in Monte Carlo simulation.

	By the properties of the probability transform, each $Z_i$ follows a uniform distribution on $[0,1]$ ($Z_i \sim Unif[0,1]$). The vector $\mathbf{Z} = (Z_1, \dots, Z_{n_Y})^T$ captures the dependence structure among the model uncertainties of $\mathbf{Y}$. The original ML model output vector $\mathbf{Y}$ can then be recovered explicitly by inverting this transformation for each component:
	\begin{equation}\label{eq:inverse_transform_Y_vector_epistemic}
		Y_i = F_{Y_i}^{-1}(Z_i;\boldsymbol{\theta}_i(\mathbf{x})) \quad \text{for } i=1, \dots, n_Y
	\end{equation}
	In vector form, this can be written as:
	\begin{equation}\label{eq:Y_from_Z_epistemic}
		\mathbf{Y} = \mathbf{F}_{\mathbf{Y}}^{-1}(\mathbf{Z};\boldsymbol{\theta}_Y(\mathbf{x}))
	\end{equation}
	
	The components of $\mathbf{Z}$ are generally dependent due to the underlying dependence structure in the joint distribution $F_{\mathbf{Y}}(\mathbf{y}; \boldsymbol{\theta})$. To characterize this dependence and obtain the joint CDF or probability distribution function (PDF) of $\mathbf{Z}$, we use the concept of a copula. By Sklar's Theorem, any multivariate joint CDF, such as the joint CDF of $\mathbf{Y}$, $F_{\mathbf{Y}}(\mathbf{y}; \boldsymbol{\theta})$, can be expressed in terms of its marginal CDFs $F_{Y_i}(y_i; \boldsymbol{\theta}_i)$ and a copula function $C_{\mathbf{Y}}$:
	\begin{equation}
		\label{eq:conditional_copula_epistemic}
		F_{\mathbf{Y}}(\mathbf{y}; \boldsymbol{\theta}) = C_{\mathbf{Y}}(F_{Y_1}(y_1; \boldsymbol{\theta}_1), \dots, F_{Y_{n_Y}}(y_{n_Y}; \boldsymbol{\theta}_{n_Y}))
	\end{equation}
	Given the definition $Z_i = F_{Y_i}(Y_i; \boldsymbol{\theta}_i)$, the joint CDF of $\mathbf{Z}$ is directly the copula function $C_{\mathbf{Y}}$:
	\begin{equation}
		\label{eq:FZ_is_copula_epistemic}
		F_{\mathbf{Z}}(z_1, \dots, z_{n_Y}) = C_{\mathbf{Y}}(z_1, \dots, z_{n_Y})
	\end{equation}
	where each $z_i \in [0,1]$. To derive the joint PDF of $\mathbf{Z}$, we differentiate its joint CDF. This gives the copula density of the distribution of $\mathbf{Y}$, denoted as $c_{\mathbf{Y}}(z_1, \dots, z_{n_Y})$:
	\begin{equation}
		\label{eq:copula_density_ZYX_epistemic}
		f_{\mathbf{Z}}(z_1, \dots, z_{n_Y}) = c_{\mathbf{Y}}(z_1, \dots, z_{n_Y}) = \frac{\partial^{n_Y} C_{\mathbf{Y}}(z_1, \dots, z_{n_Y})}{\partial z_1 \dots \partial z_{n_Y}}.
	\end{equation}
	Alternatively, using the change-of-variables formula for PDFs, if the joint PDF $f_{\mathbf{Y}}(\mathbf{y}; \boldsymbol{\theta})$ exists, and each marginal CDF $F_{Y_i}$ is differentiable and strictly increasing, then the joint PDF of $\mathbf{Z}$ is:
	\begin{multline}
		\label{eq:fZ_from_fYX_epistemic}
		f_{\mathbf{Z}}(z_1, \dots, z_{n_Y}) = f_{\mathbf{Y}}(F_{Y_1}^{-1}(z_1;\boldsymbol{\theta}_1), \dots, F_{Y_{n_Y}}^{-1}(z_{n_Y};\boldsymbol{\theta}_{n_Y})) \\
		\times \prod_{i=1}^{n_Y} \frac{1}{f_{Y_i}(F_{Y_i}^{-1}(z_i;\boldsymbol{\theta}_i);\boldsymbol{\theta}_i)}.
	\end{multline}
	This shows that the joint PDF of $\mathbf{Z}$ is the copula density, thereby encapsulating the dependence structure of $\mathbf{Y}$ while having uniform marginals.
	
	\subsubsection{Transformation of Model Uncertainty Variables (\texorpdfstring{$\mathbf{Z}$}{Z} to \texorpdfstring{$\mathbf{U}_Z$}{UZ})}
	While each $Z_i$ is uniformly distributed, the components of the vector $\mathbf{Z}$ may still be statistically dependent due to the inherent dependencies among the ML model outputs. However, some numerical algorithms require independent standard normal variables. To obtain such variables from $\mathbf{Z}$, we perform a two-step transformation:
	
	\paragraph{Step 1: Transform Uniform Variables (\texorpdfstring{$\mathbf{Z}$}{Z}) to Correlated Standard Normal Variables (\texorpdfstring{$\mathbf{W}$}{W})}
	Each component $Z_i$ is transformed into a standard normal variable $W_i$ by applying the inverse standard normal CDF, $\Phi^{-1}$:
	\begin{equation}
		\label{eq:Wi_from_Zi_epistemic}
		W_i = \Phi^{-1}(Z_i) \quad \text{for } i = 1, \dots, n_Y
	\end{equation}
	The resulting vector $\mathbf{W} = (W_1, \dots, W_{n_Y})^T$ consists of standard normal variables. $\mathbf{W}$ is a correlated standard normal vector, preserving the dependence structure of $\mathbf{Z}$ (and thus the dependence of $\mathbf{Y}$). Its joint distribution is a multivariate normal distribution $\mathcal{N}(\mathbf{0}, \boldsymbol{\Sigma_W})$, where $\boldsymbol{\Sigma_W}$ is the correlation matrix of $\mathbf{W}$. The elements of $\boldsymbol{\Sigma_W}$ are the correlation coefficients between $W_i$ and $W_j$.
	The correlation matrix $\boldsymbol{\Sigma_W}$ is generally not analytically available for arbitrary non-Gaussian distributions $F_{\mathbf{Y}}$. In such cases, it must be estimated numerically. A common approach involves generating a large number of samples of $\mathbf{Y}$ (using Monte Carlo simulation based on $F_{\mathbf{Y}}(\mathbf{y};\boldsymbol{\theta}_Y)$), transforming these $\mathbf{Y}$ samples to $\mathbf{Z}$ samples using \eqref{eq:prob_integral_transform_Z_vector_epistemic}, then to $\mathbf{W}$ samples using \eqref{eq:Wi_from_Zi_epistemic}, and finally computing the sample correlation matrix of $\mathbf{W}$ to approximate $\boldsymbol{\Sigma_W}$.
	
	\paragraph{Step 2: Decorrelate Correlated Standard Normals (\texorpdfstring{$\mathbf{W}$}{W}) to Independent Standard Normals (\texorpdfstring{$\mathbf{U}_Z$}{UZ})}
	To obtain a vector of independent standard normal variables from $\mathbf{W}$, we apply a whitening transformation using the Cholesky decomposition of its correlation matrix $\boldsymbol{\Sigma_W}$. Let $C_W$ be the lower triangular Cholesky factor of $\boldsymbol{\Sigma_W}$, such that $\boldsymbol{\Sigma_W} = C_W C_W^T$. The vector of independent standard Gaussian variables $\mathbf{U}_Z = (U_{Z,1}, \dots, U_{Z,n_Y})^T$ is then obtained by:
	\begin{equation}
		\label{eq:UZ_from_W_epistemic}
		\boxed{\mathbf{U}_Z = C_W^{-1} \mathbf{W}}
	\end{equation}
	Each component $U_{Z,k}$ is now an independent standard normal random variable. This transformation can be generally denoted as $\mathbf{U}_Z = T_Z(\mathbf{Z})$, and its inverse as $\mathbf{Z} = T_Z^{-1}(\mathbf{U}_Z)$.
	
	\subsubsection{Final Expression for Epistemic Uncertainty}
	By performing the transformations described above, we can express the ML model outputs solely as functions of independent standard normal variables. We substitute the expression for $\mathbf{Z}$ from the inverse of \eqref{eq:UZ_from_W_epistemic} (i.e., $\mathbf{Z} = \Phi(C_W\mathbf{U}_Z)$) into \eqref{eq:Y_from_Z_epistemic}:
	\begin{equation}\label{eq:Y_final_epistemic_form}
		\boxed{\mathbf{Y} = \mathbf{F}_{\mathbf{Y}}^{-1}(T_Z^{-1}(\mathbf{U}_Z);\boldsymbol{\theta}_Y)}
	\end{equation}
	This can be compactly written as:
	\begin{equation}\label{eq:Y_final_epistemic_compact}
		\boxed{\mathbf{Y} = \mathcal{G}_e(\mathbf{U}_Z)}
	\end{equation}
	where $\mathcal{G}_e$ is the composite function representing the ML models and the probabilistic transformations when only epistemic uncertainty is considered.
	
	\subsection{Case 2: General Case with Aleatory Uncertainty in \texorpdfstring{$\mathbf{X}$}{X} (X is random)}
	This section outlines the theoretical framework for quantifying system uncertainty in ML model predictions for the general case. The primary task is to determine the joint distribution of multiple ML model outputs, considering both input and model uncertainties, particularly when model uncertainties are statistically dependent.
	
	The core strategy of this framework is to transform all sources of uncertainty---both input uncertainties ($\mathbf{X}$) and model uncertainties (represented by an auxiliary vector $\mathbf{Z}$)---into a unified space of independent standard normal variables. This is particularly crucial because the parameters governing the model uncertainty (e.g., means and variances of ML predictions) are often functions of the random inputs $\mathbf{X}$, leading to a strong coupling between these two types of uncertainty. This decoupling significantly simplifies subsequent uncertainty propagation and analysis.
	
	\paragraph{Transformation of Input Variables (\texorpdfstring{$\mathbf{X}$}{X} to \texorpdfstring{$\mathbf{U}_X$}{UX})}
	The input vector $\mathbf{X}$ (size $n_X$) can consist of random variables that may be statistically dependent and follow arbitrary marginal distributions. We transform $\mathbf{X}$ into a vector of independent standard normal variables, $\mathbf{U}_X = (U_{X,1}, \dots, U_{X,n_X})^T$. This transformation is a standard procedure in physics-based reliability analysis, often achieved through methods like the Nataf transformation or sequential inverse CDF transformations such as Rosenblatt Transformation. Such a transformation allows us to express $\mathbf{X}$ as a function of $\mathbf{U}_X$:
	\begin{equation}\label{eq:X_from_UX_general}
		\mathbf{X} = T_X(\mathbf{U}_X)
	\end{equation}
	where $T_X$ is the transformation function. Each component $U_{X,j}$ is an independent standard normal random variable.
	
	\paragraph{Decoupling Model Uncertainty via \texorpdfstring{$\mathbf{Z}$}{Z}}
	For each ML model output $Y_i$ in the vector $\mathbf{Y}$, its probabilistic response given $\mathbf{X}$ is characterized by a conditional distribution $F_{Y_i|\mathbf{X}}(Y_i; \boldsymbol{\theta}_i(\mathbf{X}))$. The auxiliary scalar random variable $Z_i$ for each $Y_i$ is defined by
	\begin{equation}\label{eq:prob_integral_transform_Z_vector_general}
		Z_i = F_{Y_i|\mathbf{X}}(Y_i;\boldsymbol{\theta}_i(\mathbf{X})) \quad \text{for } i=1, \dots, n_Y
	\end{equation}
	As discussed previously, each $Z_i$ follows a uniform distribution on $[0,1]$ ($Z_i \sim Unif[0,1]$) and is independent of $\mathbf{X}$. The vector $\mathbf{Z} = (Z_1, \dots, Z_{n_Y})^T$ captures the conditional dependence structure among the model uncertainties of $\mathbf{Y}$ given $\mathbf{X}$. The original ML model output vector $\mathbf{Y}$ can then be recovered by inverting this transformation for each component:
	\begin{equation}\label{eq:inverse_transform_Y_vector_general}
		Y_i = F_{Y_i|\mathbf{X}}^{-1}(Z_i;\boldsymbol{\theta}_i(\mathbf{X})) \quad \text{for } i=1, \dots, n_Y
	\end{equation}
	In vector form, this can be written as:
	\begin{equation}\label{eq:Y_from_Z_and_X_general}
		\mathbf{Y} = \mathbf{F}_{\mathbf{Y}|\mathbf{X}}^{-1}(\mathbf{Z};\boldsymbol{\theta}_Y(\mathbf{X}))
	\end{equation}
	This equation explicitly shows that $\mathbf{Y}$ is a function of the random input variables $\mathbf{X}$ and the model uncertainty variables $\mathbf{Z}$, which are now decoupled. The components of $\mathbf{Z}$ are generally dependent due to the underlying dependence structure in the conditional joint distribution $F_{\mathbf{Y}|\mathbf{X}}(\mathbf{y}; \boldsymbol{\theta}(\mathbf{X}))$. The multivariate joint CDF of $\mathbf{Y}$ given $\mathbf{X}$, $F_{\mathbf{Y}|\mathbf{X}}(\mathbf{y}; \boldsymbol{\theta}(\mathbf{X}))$, can be expressed in terms of its conditional marginal CDFs $F_{Y_i|\mathbf{X}}(y_i; \boldsymbol{\theta}_i(\mathbf{X}))$ and a copula function $C_{\mathbf{Y}|\mathbf{X}}$:
	\begin{equation}
		\label{eq:conditional_copula_general}
		F_{\mathbf{Y}|\mathbf{X}}(\mathbf{y}; \boldsymbol{\theta}(\mathbf{X})) = C_{\mathbf{Y}|\mathbf{X}}(F_{Y_1|\mathbf{X}}(y_1; \boldsymbol{\theta}_1(\mathbf{X})), \dots, F_{Y_{n_Y}|\mathbf{X}}(y_{n_Y}; \boldsymbol{\theta}_{n_Y}(\mathbf{X})))
	\end{equation}
	Given the definition $Z_i = F_{Y_i|\mathbf{X}}(Y_i; \boldsymbol{\theta}_i(\mathbf{X}))$, the joint CDF of $\mathbf{Z}$ is directly the copula function $C_{\mathbf{Y}|\mathbf{X}}$:
	\begin{equation}
		\label{eq:FZ_is_copula_general}
		F_{\mathbf{Z}}(z_1, \dots, z_{n_Y}) = C_{\mathbf{Y}|\mathbf{X}}(z_1, \dots, z_{n_Y})
	\end{equation}
	where each $z_i \in [0,1]$. To derive the joint PDF of $\mathbf{Z}$, we differentiate its joint CDF. This gives the copula density of the conditional distribution of $\mathbf{Y}$ given $\mathbf{X}$, denoted as $c_{\mathbf{Y}|\mathbf{X}}(z_1, \dots, z_{n_Y})$:
	\begin{equation}
		\label{eq:copula_density_ZYX_general}
		f_{\mathbf{Z}}(z_1, \dots, z_{n_Y}) = c_{\mathbf{Y}|\mathbf{X}}(z_1, \dots, z_{n_Y}) = \frac{\partial^{n_Y} C_{\mathbf{Y}|\mathbf{X}}(z_1, \dots, z_{n_Y})}{\partial z_1 \dots \partial z_{n_Y}}.
	\end{equation}
	Alternatively, using the change-of-variables formula for PDFs, if the conditional joint PDF $f_{\mathbf{Y}|\mathbf{X}}(\mathbf{y}; \boldsymbol{\theta}(\mathbf{X}))$ exists, and each conditional marginal CDF $F_{Y_i|\mathbf{X}}$ is differentiable and strictly increasing, then the joint PDF of $\mathbf{Z}$ is:
	\begin{multline}
		\label{eq:fZ_from_fYX_general}
		f_{\mathbf{Z}}(z_1, \dots, z_{n_Y}) = f_{\mathbf{Y}|\mathbf{X}}(F_{Y_1|\mathbf{X}}^{-1}(z_1;\boldsymbol{\theta}_1(\mathbf{X})), \dots, F_{Y_{n_Y}|\mathbf{X}}^{-1}(z_{n_Y};\boldsymbol{\theta}_{n_Y}(\mathbf{X}))) \\
		\times \prod_{i=1}^{n_Y} \frac{1}{f_{Y_i|\mathbf{X}}(F_{Y_i|\mathbf{X}}^{-1}(z_i;\boldsymbol{\theta}_i(\mathbf{X}));\boldsymbol{\theta}_i(\mathbf{X}))}.
	\end{multline}
	This shows that the joint PDF of $\mathbf{Z}$ is the copula density, thereby encapsulating the conditional dependence structure of $\mathbf{Y}$ given $\mathbf{X}$ while having uniform marginals.
	
	\paragraph{Transformation of Model Uncertainty Variables (\texorpdfstring{$\mathbf{Z}$}{Z} to \texorpdfstring{$\mathbf{U}_Z$}{UZ})}
	While each $Z_i$ is uniformly distributed and independent of $\mathbf{X}$, the components of the vector $\mathbf{Z}$ may still be statistically dependent due to the inherent dependencies among the ML model outputs. To obtain independent standard normal variables from $\mathbf{Z}$, we perform the same two-step transformation as described in Sec. 3.2. The final result is
	
	\begin{equation}
		\label{eq:UZ_from_W_general}
		\boxed{\mathbf{U}_Z = C_W(\mathbf{X})^{-1} \mathbf{W}}
	\end{equation}
	Each component $U_{Z,k}$ is now an independent standard normal random variable. This transformation can be generally denoted as $\mathbf{U}_Z = T_Z(\mathbf{Z};\mathbf{X})$, and its inverse as $\mathbf{Z} = T_Z^{-1}(\mathbf{U}_Z;\mathbf{X})$.
	
	\subsubsection{Unified Independent Space and Final Expression}
	By performing the transformations described above, we can express all sources of uncertainty in a unified space of independent standard normal variables. We combine the transformed input variables $\mathbf{U}_X$ and the transformed model uncertainty variables $\mathbf{U}_Z$ into a single comprehensive vector $\mathbf{U}$:
	\begin{equation}\label{eq:U_total_general}
		\mathbf{U} = (\mathbf{U}_X, \mathbf{U}_Z)
	\end{equation}
	All components of $\mathbf{U}$ are now independent standard normal random variables. Finally, we substitute the expressions for $\mathbf{X}$ from \eqref{eq:X_from_UX_general} and $\mathbf{Z}$ from the inverse of \eqref{eq:UZ_from_W_general} (i.e., $\mathbf{Z} = \Phi(C_W(\mathbf{X})\mathbf{U}_Z)$) into \eqref{eq:Y_from_Z_and_X_general}:
	\begin{equation}\label{eq:Y_final_general_form_overall}
		\boxed{\mathbf{Y} = \mathbf{F}_{\mathbf{Y}|T_X(\mathbf{U}_X)}^{-1}(T_Z^{-1}(\mathbf{U}_Z; T_X(\mathbf{U}_X));\boldsymbol{\theta}_Y(T_X(\mathbf{U}_X)))}
	\end{equation}
	This can be compactly written as:
	\begin{equation}\label{eq:Y_final_compact_overall}
		\boxed{\mathbf{Y} = \mathcal{G}(\mathbf{U}_X, \mathbf{U}_Z) = \mathcal{G}(\mathbf{U})}
	\end{equation}
	where $\mathcal{G}$ is the composite function representing the ML models and all the probabilistic transformations.
	
	\paragraph{Advantages of Transformation to Independent Standard Normal Variables}
	
	This transformation offers significant advantages for UQ:
	\begin{itemize}
		\item \textbf{Decoupling of Uncertainties:} The primary benefit is the explicit decoupling of input uncertainty and model uncertainty. This allows for a clear separation and independent treatment of aleatory and epistemic uncertainties, even when they are originally coupled.
		\item \textbf{Simplified Uncertainty Propagation:} By expressing $\mathbf{Y}$ as a function of only independent standard normal variables $\mathbf{U}$, subsequent uncertainty propagation (e.g., moment methods, fast probability integration, or other advanced UQ techniques) becomes significantly simpler and more computationally efficient. Many UQ algorithms are optimized for such inputs; for example, the intergrand for the fast probability integration becomes symmetric multivariate standard normal PDF.
		\item \textbf{Facilitated Sensitivity Analysis:} This framework naturally lends itself to sensitivity analysis. Techniques like Sobol' indices, which require independent input variables, can be directly applied to $\mathcal{G}(\mathbf{U})$ to quantify the contribution of each independent uncertainty source (i.e., each component of $\mathbf{U}$) to the total variance of $\mathbf{Y}$. This provides valuable insights into the most influential sources of uncertainty.
	\end{itemize}
	
	\section{GP Models}
	In this section, we apply the framework developed in Sec. 3.3 to GP models. For a multi-output GP model, the prediction at a new input $\mathbf{X}$ (test point) is a vector of random variables, $\mathbf{Y} = (Y_1, \dots, Y_{n_Y})^T$, whose joint distribution is a multivariate Gaussian. This is expressed as:
	$$
	\mathbf{Y} | \mathbf{X} \sim \mathcal{N}(\boldsymbol{\mu}(\mathbf{X}), \boldsymbol{\Sigma}(\mathbf{X}))
	$$
	Here, $\boldsymbol{\mu}(\mathbf{X})$ is the vector of predictive means and $\boldsymbol{\Sigma}(\mathbf{X})$ is the predictive covariance matrix, both of which are functions of the input vector $\mathbf{X}$. The parameters of this conditional distribution are given by $\boldsymbol{\theta}_Y(\mathbf{X}) = \{\boldsymbol{\mu}(\mathbf{X}), \boldsymbol{\Sigma}(\mathbf{X})\}$.
	
	Following the methodology in Sec. 3.3, we can derive the necessary equations for the GP case.
	
	\paragraph{Transformation to Independent Standard Normal Variables}
	The core of the methodology is to transform the dependent input random variables into a set of independent standard normal variables.
	
	\subparagraph{Step 1: Transform Correlated Gaussian Variables (\texorpdfstring{$\mathbf{Y}$}{Y}) to Independent Standard Normals (\texorpdfstring{$\mathbf{U}_Y$}{UY})}
	Given that $\mathbf{Y}$ is a multivariate normal vector conditional on $\mathbf{X} = \mathbf{x}$, the transformation to independent standard normal variables is straightforward and analytical. We perform a whitening transformation using the Cholesky decomposition of the conditional covariance matrix $\boldsymbol{\Sigma}(\mathbf{x})$. Let $C_Y(\mathbf{x})$ be the lower triangular Cholesky factor of $\boldsymbol{\Sigma}(\mathbf{x})$, such that $\boldsymbol{\Sigma}(\mathbf{x}) = C_Y(\mathbf{x}) C_Y(\mathbf{x})^T$. The vector of independent standard Gaussian variables $\mathbf{U}_Y = (U_{Y,1}, \dots, U_{Y,n_Y})^T$ is then obtained by:
	\begin{equation}
		\label{eq:UY_from_Y}
		\mathbf{U}_Y = C_Y(\mathbf{X})^{-1} (\mathbf{Y} - \boldsymbol{\mu}(\mathbf{x}))
	\end{equation}
	Each component $U_{Y,k}$ is now an independent standard normal random variable.
	
	\subparagraph{Step 2: Express \texorpdfstring{$\mathbf{Y}$}{Y} as a Function of \texorpdfstring{$\mathbf{U}_Y$}{UY}}
	We can express the original ML prediction vector $\mathbf{Y}$ as a function of the independent standard normal variables $\mathbf{U}_Y$ by inverting the transformation in \eqref{eq:UY_from_Y}:
	\begin{equation}
		\label{eq:Y_from_UY}
		\boxed{\mathbf{Y} = \boldsymbol{\mu}(\mathbf{X}) + C_Y(\mathbf{X})\mathbf{U}_Y}
	\end{equation}
	This equation is analogous to \eqref{eq:Y_final_epistemic_compact}, where the transformations $T_Z$ and $T_Z^{-1}$ simplify significantly due to the Gaussian nature of the GP predictions.
	
	\paragraph{Unified Independent Space}
	For the general case with random inputs $\mathbf{X}$, we use the same strategy as in Sec. 3.3. We have the transformation for the input variables:
	\begin{equation}
		\mathbf{X} = T_X(\mathbf{U}_X)
	\end{equation}
	and the relationship for the GP outputs derived above, which is now a function of the random input $\mathbf{X}$:
	\begin{equation}
		\mathbf{Y} = \boldsymbol{\mu}(\mathbf{X}) + C_Y(\mathbf{X})\mathbf{U}_Y
	\end{equation}
	We define the total vector of independent standard normal variables as $\mathbf{U} = (\mathbf{U}_X, \mathbf{U}_Y)^T$. Substituting the expression for $\mathbf{X}$ into the equation for $\mathbf{Y}$, we get the final form analogous to \eqref{eq:Y_final_compact_overall}:
	\begin{equation}
		\label{eq:Y_final_GP_form}
		\boxed{\mathbf{Y} = \boldsymbol{\mu}(T_X(\mathbf{U}_X)) + C_Y(T_X(\mathbf{U}_X))\mathbf{U}_Y}
	\end{equation}
	This can be written compactly as $\mathbf{Y} = \mathcal{G}(\mathbf{U})$, where all components of $\mathbf{U}$ are independent standard normal random variables. This result simplifies the general framework for GP models because the probability integral transform and its inverse steps (represented by $T_Z$ and $T_Z^{-1}$ in the general case) are naturally replaced by the simpler Cholesky transformation due to the Gaussian assumption.
	
	\section{Numerical Algorithm Development}
	
	Numerical algorithms can be developed based on the proposed theoretical framework. Accurate numerical algorithms are essential for system UQ. Without it, the benefit of accounting for ML model uncertainty is greatly diminished. Challenges must be addressed for the development.
	
	The primary issue is that existing fast probability integration methods, while effective for some problems, can introduce substantial errors. This is particularly true when applied to the expanded uncertainty space that includes model uncertainty. These errors are caused by two main factors. First, the integration boundaries, determined by equations such as (15), (23), and (28), may be highly irregular, leading to numerical instability. This is caused by model uncertainty that varies in the space of random input variables. Second, the ML model's nonlinearity, when coupled with this irregular boundary, can result in large integration errors, which compromise the benefits of propagating model uncertainty. To overcome these challenges, it is important to carefully and deliberately treat the model uncertainty variable to reduce its impact and improve the accuracy of the numerical integration.
	
	Handling high-dimensional dependent model outputs in system UQ is a nontrivial task. Numerical methods for integrating high-dimensional joint PDF are currently limited, especially when dependencies exist among model responses. For example, computing the joint probability or CDF of a multivariate normal distribution typically relies on sampling-based methods when the dimensionality exceeds three. These methods often lack accuracy in estimating rare failure events due to the curse of dimensionality and insufficient sampling in critical regions \cite{Yin2021}. As a result, efficient and accurate numerical integration techniques are urgently needed for problems involving a large number of statistically dependent outputs.
	
	Alternatively, the Monte Carlo method can be used directly with decoupled uncertainty. Although it is straightforward to use the MC method, it is inefficient for rare joint probabilities in the range between $10^{-6}$ and $10^{-9}$. The Monte Carlo method may also cause numerical problems when used in optimization, such as reliability-based design optimization. Therefore, numerical methods are preferred over sampling methods for practical engineering applications.

	\section{Conclusions}
	This paper presents a theoretical framework that provides essential formulations for developing numerical algorithms to quantify system uncertainty. The core of this framework is its ability to decouple two distinct types of uncertainty: the randomness in input variables and the inherent uncertainty of machine learning models. This decoupling is achieved through specific formulations that expand the uncertainty space and establish new formulations, as defined by equations such as (15), (23), and (28). These foundational steps are critical for accurate uncertainty quantification that accounts for the complete range of system-level variables. By providing these key formulations, our framework directly supports the development of new numerical algorithms tailored to address the challenges of model uncertainty. It highlights that the successful implementation of such algorithms relies on a deliberate approach to manage the irregular integration boundaries and the nonlinearity of ML models, which are often the primary sources of numerical instability. This work therefore serves as a vital blueprint for future research aimed at creating efficient and reliable numerical methods that are essential for practical engineering applications, especially those that involve rare joint probabilities.

	\bibliographystyle{unsrt}
	\bibliography{references}		
\end{document}